\documentclass[11pt,letterpaper]{article}
\usepackage{emnlp2017}
\usepackage{times}
\usepackage{latexsym}
\usepackage{graphicx}
\usepackage{amssymb}
\usepackage{multirow}
\usepackage{amsmath}
\usepackage{mathtools}
\usepackage{tabularx}
\usepackage{color}
\usepackage{tablefootnote}

\newcommand{\argmax}{\arg\!\max}

\aclfinalcopy

\title{Neural Transition-based Syntactic Linearization}

\author{Linfeng Song$^1$, Yue Zhang$^2$ \and Daniel Gildea$^1$ \\
  $^1$Department of Computer Science, University of Rochester, Rochester, NY 14627 \\
  $^2$School of Engineering, Westlake University, China}

\date{}

\begin{document}

\maketitle

\begin{abstract}
  The task of linearization is to find a grammatical order given a set of words.
  Traditional models use statistical methods.
  Syntactic linearization systems, which generate a sentence along with its syntactic tree, have shown state-of-the-art performance.
  Recent work shows that a multi-layer LSTM language model outperforms competitive statistical syntactic linearization systems without using syntax.
  In this paper, we study neural syntactic linearization, building a transition-based syntactic linearizer leveraging a feed forward neural network, observing significantly better results compared to LSTM language models on this task.
\end{abstract}

\section{Introduction}

\emph{Linearization} is the task of finding the grammatical order for a given set of words.
Syntactic linearization systems generate output sentences along with their syntactic trees. 
Depending on how much syntactic information is available during decoding, recent work on syntactic linearization can be classified into 
abstract word ordering \cite{wan-EtAl:2009:EACL,zhang-blackwood-clark:2012:EACL2012,degispert-tomalin-byrne:2014:EACL}, where no syntactic information is available during decoding,
full tree linearization \cite{he-EtAl:2009:ACLIJCNLP,bohnet-EtAl:2010:PAPERS,song2014joint}, where full tree information is available,
and partial tree linearization \cite{zhang2013partial}, where partial syntactic information is given as input.
Linearization has been adapted to tasks such as machine translation \cite{zhang-EtAl:2014:EMNLP20141}, 
and is potentially helpful for many NLG applications, such as cooking recipe generation \cite{kiddon-zettlemoyer-choi:2016:EMNLP2016}, dialogue response generation \cite{wen-EtAl:2015:EMNLP}, and question generation \cite{serban-EtAl:2016:P16-1}.

Previous work \cite{wan-EtAl:2009:EACL,liu-EtAl:2015:NAACL-HLT1} has shown that jointly predicting the syntactic tree and the surface string gives better results by allowing syntactic information to guide statistical linearization.
On the other hand, most such methods employ statistical models with discriminative features.
Recently, \newcite{schmaltz-rush-shieber:2016:EMNLP2016} report new state-of-the-art results by leveraging a neural language model \emph{without} using syntactic information.
In their experiments, the neural language model, which is less sparse and captures long-range dependencies, outperforms previous discrete syntactic systems.

A research question that naturally arises from this result is whether syntactic information is helpful for a \emph{neural} linearization system.
We empirically answer this question by comparing a neural transition-based syntactic linearizer with the neural language model of \newcite{schmaltz-rush-shieber:2016:EMNLP2016}.
Following \newcite{liu-EtAl:2015:NAACL-HLT1}, our linearizer works incrementally given a set of words, using a stack to store partially built dependency trees, and a set to maintain \emph{unordered} incoming words.
At each step, it either \emph{shifts} a word onto the stack, or \emph{reduces} the top two partial trees on the stack.
We leverage a feed forward neural network, which takes stack features as input and predicts the next action (such as \textsc{Shift}, \textsc{LeftArc} and \textsc{RightArc}).
Hence our method can be regarded as an extension of the parser of \newcite{chen-manning:2014:EMNLP2014}, adding word ordering functionalities.

In addition, we investigate two methods for integrating neural language models: 
interpolating the log probabilities of both models 
and integrating the neural language model as a feature.
On standard benchmarks, our syntactic linearizer gives results that are higher than the LSTM language model of \newcite{schmaltz-rush-shieber:2016:EMNLP2016} by 7 BLEU points \cite{papineni-EtAl:2002:ACL} using greedy search, 
and the gap can go up to 11 BLEU points by integrating the LSTM language model as features.
The integrated system also outperforms the LSTM language model by 1 BLEU point using  beam search,
which shows that syntactic information is useful for a neural linearization system.

\section{Related work}

Previous work \cite{white:2005:Software,white-rajkumar:2009:EMNLP,zhang-clark:2011:EMNLP,zhang2013partial} on syntactic linearization uses best-first search, which adopts a priority queue to store partial hypotheses and a chart to store input words.
At each step, it pops the highest-scored hypothesis from the priority queue, expanding it by combination with the words in the chart, before finally putting all new hypotheses back into the priority queue.
As the search space is huge, a timeout threshold is set, beyond which the search terminates and the current best hypothesis is taken as the result.

\newcite{liu-EtAl:2015:NAACL-HLT1} adapt the transition-based dependency parsing algorithm for the linearization task by allowing the transition-based system to shift any word in the given set, rather than the first word in the buffer as in dependency parsing.
Their results show much lower search times and higher performance compared to \newcite{zhang2013partial}.
Following this line, \newcite{liu-zhang:2015:EMNLP} further improve the performance by incorporating an $n$-gram language model.
Our work takes the transition-based framework, but is different in two main aspects: first, we train a feed-forward neural network for making decisions, while they all use perceptron-like models.
Second, we investigate a light version of the system, which only uses word features,
while previous works all rely on POS tags and arc labels, limiting their usability on low-resource domains and languages.

\newcite{schmaltz-rush-shieber:2016:EMNLP2016} are the first to adopt neural networks on this task, while only using \emph{surface} features.
To our knowledge, we are the first to leverage both neural networks and \emph{syntactic} features.
The contrast between our method and the method of \newcite{chen-manning:2014:EMNLP2014} is reminiscent of the contrast between the method of \newcite{liu-EtAl:2015:NAACL-HLT1} and the dependency parser of \newcite{zhang-nivre:2011:ACL-HLT2011}.
Comparing with the dependency parsing task, which assumes that POS tags are available as input, the search space of syntactic linearization is much larger.

Recent work \citep{zhang2013partial,song2014joint,liu-EtAl:2015:NAACL-HLT1,liu-zhang:2015:EMNLP} on syntactic linearization uses dependency grammar.
We follow this line of works.
On the other hand, linearization with other syntactic grammars, such as context free grammar \cite{degispert-tomalin-byrne:2014:EACL} and combinatory categorial grammar \cite{white-rajkumar:2009:EMNLP,zhang-clark:2011:EMNLP}, has also been studied.

\section{Task}

Given an input bag-of-words $x=\{x_1,x_2,...,x_n\}$, the goal is to output the correct permutation $y$, which recovers the original sentence, from the set of all possible permutations $\mathcal{Y}$.
A linearizer can be seen as a scoring function $f$ over $\mathcal{Y}$, which is trained to output its highest scoring permutation $\hat{y}=\argmax_{y'\in \mathcal{Y}}f(x,y')$ as close as possible to the correct permutation $y$.

\subsection{Baseline: an LSTM language model}

The LSTM language model of \newcite{schmaltz-rush-shieber:2016:EMNLP2016} is similar to the medium LSTM setup of \newcite{zaremba2014recurrent}.
It contains two LSTM layers, each of which has 650 hidden units and is followed by a dropout layer during training.
The multi-layer LSTM language model can be represented as:
\begin{align}
\mathbf{h}_{t,i}, \mathbf{c}_{t,i} = \mathrm{LSTM}(\mathbf{h}_{t,i-1},\mathbf{h}_{t-1,i},\mathbf{c}_{t-1,i}) \\
p(w_{t,j}|w_{t-1},...,w_1) = \frac{\exp(\mathbf{v}_j^\intercal \mathbf{h}_{t,I})}
{\sum_{j'}\exp(\mathbf{v}_{j'}^\intercal \mathbf{h}_{t,I})}\textrm{,}
\end{align}
where $\mathbf{h}_{t,i}$ and $\mathbf{c}_{t,i}$ are the output and cell memory of the $i$-th layer at step $t$, respectively, $\mathbf{h}_{t,0}=\mathbf{x}_t$ is the input of the network at step $t$, $I$ is the number of layers, $w_{t,j}$ represents outputting $w_j$ at $t$ step, $\mathbf{v}_j$ is the embedding of $w_j$,
and the LSTM function is defined as:
\begin{align}
\left (
  \begin{tabular}{c}
  $\mathbf{i}$ \\
  $\mathbf{f}$ \\
  $\mathbf{o}$ \\ 
  $\mathbf{g}$ \\
  \end{tabular}
\right ) &= 
\left (
  \begin{tabular}{c}
  $\sigma$ \\
  $\sigma$ \\
  $\sigma$ \\ 
  $\mathrm{tanh}$ \\
  \end{tabular}
\right ) \mathbf{W}_{4n,2n}
\left (
  \begin{tabular}{c}
  $\mathbf{h}_{t,i-1}$ \\
  $\mathbf{h}_{t-1,i}$ \\
  \end{tabular}
\right ) \\
\mathbf{c}_{t,i} &= \mathbf{f}\odot \mathbf{c}_{t-1,i}+\mathbf{i}\odot \mathbf{g} \\
\mathbf{h}_{t,i} &= \mathbf{o}\odot \mathrm{tanh}(\mathbf{c}_{t,i})\textrm{,}
\end{align}
where $\sigma$ is the sigmoid function, $\mathbf{W}_{4n,2n}$ is the weights of LSTM cells, and $\odot$ is the element-wise product operator.

\begin{figure}
\centering
\includegraphics[scale=0.8]{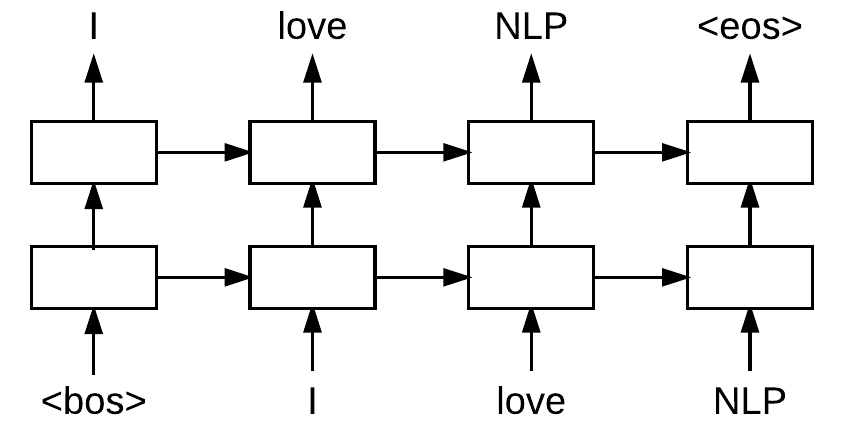}
\caption{Linearization procedure of the baseline.}
\label{fig:baseline}
\end{figure}

Figure \ref{fig:baseline} shows the linearization procedure of the baseline system, when taking the bag-of-words \{``NLP'',``love'',``I''\} as input.
At each step, it takes the output word from the previous step as input and predicts the current word, which is chosen from the remaining input bag-of-words rather than from the entire vocabulary.
Therefore it takes $n$ steps to linearize a input consisting of $n$ words.

\section{Neural transition-based syntactic linearization}

Transition-based syntactic linearization can be considered as an extension to transition-based dependency parsing \cite{liu-EtAl:2015:NAACL-HLT1}, with the main difference being that the word order is not given in the input, so that any word can be shifted at each step.
This leads to a much larger search space.
In addition, under our setting, \emph{extra} dependency relations or POS on input words are not available.

\begin{figure}
\centering
\includegraphics[scale=0.9]{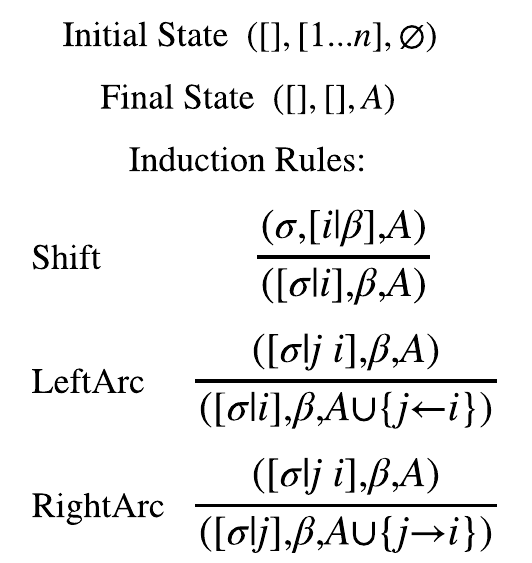}
\caption{Deduction system of transition-based syntactic linearization}
\label{fig:trans_syn_lin}
\end{figure}

The output building process is modeled as a state-transition process.
As shown in Figure \ref{fig:trans_syn_lin}, each state $s$ is defined as $(\sigma, \rho, A)$, where $\sigma$ is a stack that maintains a partial derivation, $\rho$ is an \emph{unordered} set of incoming input words and $A$ is the set of dependency relations that have been built.
Initially, the stack $\sigma$ is empty, while the set $\rho$ contains all the input words, and the set of dependency relations $A$ is empty.
At the end, the set $\rho$ is empty, while $A$ contains all dependency relations for the predicted dependency tree.
At a certain state, 
a \textsc{Shift} action chooses one word from the set $\rho$ and pushes it onto the stack $\sigma$, 
a \textsc{LeftArc} action makes a new arc $\{j \leftarrow i\}$ from the stack's top two items ($i$ and $j$), 
while a \textsc{RightArc} action makes a new arc $\{j \rightarrow i\}$ from $i$ and $j$.
Using these possible actions, the unordered word set \{``NLP$_0$'',``love$_1$'',``I$_2$''\} is linearized as shown in Table \ref{tab:transition}, and the result is ``I$_2$ $\leftarrow$ love$_1$ $\rightarrow$ NLP$_0$''.\footnote{For a clearer introduction to our state-transition process, we omit the \textsc{Pos}-$p$ actions, which are introduced in Section \ref{sec:actions}. In our implementation, each \textsc{Shift}-$w$ is followed by exact one \textsc{Pos}-$p$ action.}

\begin{table} \small
\centering
\begin{tabular}{lllll}
\hline
step & action & $\sigma$ & $\rho$  & $A$ \\
\hline
\hline
init &              & $[]$      & $(1~2~3)$ & $\varnothing$ \\
0    & {Shift-I}    & $[1]$     & $(2~3)$   &  \\
1    & {Shift-love} & $[1~2]$   & $(3)$     &  \\
2    & {Shift-NLP}  & $[1~2~3]$ & $()$      & \\
3    & {RArc-dobj}  & $[1~2]$   & $()$      & $A \cup \{2 \rightarrow 3\}$  \\
4    & {LArc-nsubj} & $[2]$     & $()$      & $A \cup \{1 \leftarrow 2\}$ \\
5    & {End}        & $[]$      & $()$      & $A$ \\
\hline
\end{tabular}
\caption{Transition-based syntactic linearization for ordering \{``NLP$_3$'',``love$_2$'',``I$_1$''\}, where \emph{RArc} and \emph{LArc} are the abbreviations for RightArc and LeftArc, respectively. More details on actions are in Section \ref{sec:actions}.}
\label{tab:transition}
\end{table}

\subsection{Model}
\label{sec:model}

\begin{figure}[t]
\includegraphics[scale=.8]{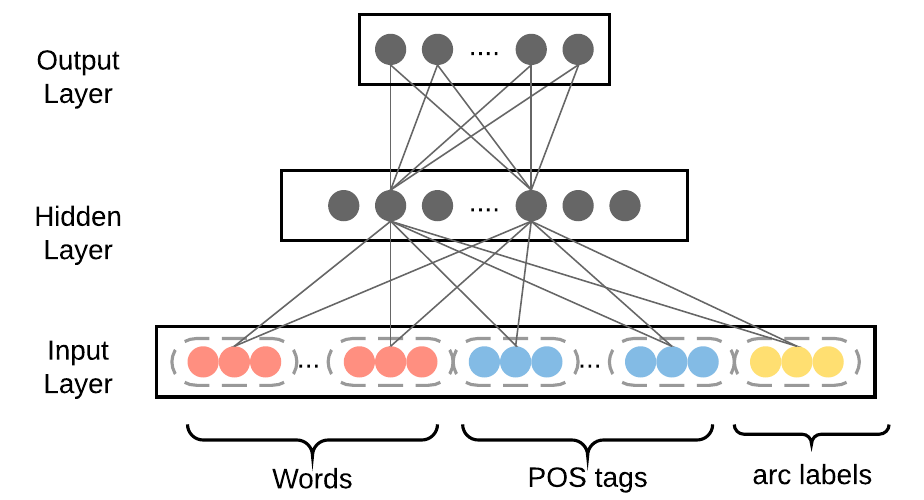}
\caption{Neural syntactic linearization model}
\label{fig:sys}
\end{figure}

To predict the next transition action for a given state, our linearizer makes use of a feed-forward neural network to score the actions as shown in Figure \ref{fig:sys}.
The network takes a set of word, POS tag, and arc label features from the stack as input and outputs the probability distribution of the next actions.
In particular, we represent each word as a $d$-dimensional vector $\mathbf{e}_i^w \in \mathbb{R}^d$ using a word embedding matrix is $\mathbf{E}^w \in \mathbb{R}^{d\times N_w}$, where $N_w$ is the vocabulary size. 
Similarly each POS tag and arc label are also mapped to a $d$-dimensional vector, where $\mathbf{e}_j^t,\mathbf{e}_k^l \in \mathbb{R}^d$ are the representations of the $j$-th POS tag and $k$-th arc label, respectively.
The embedding matrices of POS tags and arc labels are $\mathbf{E}^t \in \mathbb{R}^{d\times N_t}$ and $\mathbf{E}^l \in \mathbb{R}^{d\times N_l}$, where $N_t$ and $N_l$ correspond to the number of POS tags and arc labels, respectively.
We choose a set of feature words, POS tags, and arc labels from the stack context, using their embeddings as input to our neural network.
Next, we map the input layer to the hidden layer via:
\begin{equation}
h=g(\mathbf{W}_1^w \mathbf{x}^w+\mathbf{W}_1^t \mathbf{x}^t + \mathbf{W}_1^l \mathbf{x}^l + \mathbf{b}_1)\textrm{,}
\end{equation}
where $\mathbf{x}^w$, $\mathbf{x}^t$, and $\mathbf{x}^l$ are the concatenated feature word embeddings, POS tag embeddings, and arc label embeddings, respectively, $\mathbf{W}_1^w$, $\mathbf{W}_1^t$, and $\mathbf{W}_1^l$ are the corresponding weight matrices, $\mathbf{b}_1$ is the bias term and $g()$ is the activation function of the hidden layer.
The word, POS tag and arc label features are described in Section \ref{sec:features}.

Finally, the hidden vector $\mathbf{h}$ is mapped to an output layer, which uses a softmax activation function for modeling multi-class action probabilities:
\begin{equation}
p(a|s,\theta)=\mathrm{softmax}(\mathbf{W}_2\mathbf{h})\textrm{,}
\end{equation}
where $p(a|s,\theta)$ represents the probability distribution of the next action.
There is no bias term in this layer and the model parameter $\mathbf{W}_2$ can also be seen as the embedding matrix of all actions.

\subsection{Actions}
\label{sec:actions}

We use 5 types of actions: 
\begin{itemize}
\item \textsc{Shift}-$w$ pushes a word $w$ onto the stack.
\item \textsc{Pos}-$p$ assigns a POS tag $p$ to the newly shifted word.
\item \textsc{LeftArc}-$l$ pops the top two items $i$ and $j$ off stack and pushes $\{j \xleftarrow[]{\text{$l$}} i\}$ onto the stack.
\item \textsc{RightArc}-$l$ pops the top two items $i$ and $j$ off stack and pushes $\{j \xrightarrow[]{\text{$l$}} i\}$ onto the stack.
\item \textsc{End} ends the decoding procedure.
\end{itemize}
Given a set of $n$ words as input, the linearizer takes $3n$ steps to synthesize the sentence.
The number of actions is large, making it computationally inefficient to do softmax over all actions.
Here for each set of words $S$ we only consider all possible actions for linearizing the set, which constraints \textsc{Shift}-$w_i$ to all words in the set.

\begin{table} \small
\centering
\begin{tabular}{l|l} 
\hline
(1) & $S_1.w$; $S_1.t$; $S_2.w$; $S_2.t$; $S_3.w$; $S_3.t$;  \\
\hline
\multirow{5}{1em}{(2)} & $i = 1,2$ \\
					   & ~~~$lc_1(S_i).w$; $lc_1(S_i).t$; $lc_1(S_i).l$; \\
                       & ~~~$lc_2(S_i).w$; $lc_2(S_i).t$; $lc_2(S_i).l$; \\
                       & ~~~$rc_1(S_i).w$; $rc_1(S_i).t$; $rc_1(S_i).l$; \\
                       & ~~~$rc_2(S_i).w$; $rc_2(S_i).t$; $rc_2(S_i).l$; \\

\hline
\multirow{4}{1em}{(3)} & $i = 1,2$ \\
					   & ~~~$lc_1(lc_1(S_i)).w$; $lc_1(lc_1(S_i)).t$; \\
                       & ~~~$lc_1(lc_1(S_i)).l$; $rc_1(rc_1(S_i)).w$; \\
                       & ~~~$rc_1(rc_1(S_i)).t$; $rc_1(rc_1(S_i)).l$; \\
\hline
\end{tabular}
\caption{Feature templates, where $S_i$ denotes the $i$th item on the stack, $w$, $t$ and $l$ denotes the word, POS tag and arc label, respectively.}
\label{tab:feat}
\end{table}

\subsection{Features}
\label{sec:features}

The feature templates our model uses are shown in Table \ref{tab:feat}. We pick (1) the words and POS tags of the top 3 items on the stack, (2) the words, POS tags, and arc labels of the first and the second leftmost / rightmost children of the top 2 items on the stack and (3) the words, POS tags and arc labels of the leftmost of leftmost / rightmost of rightmost children of the top two items on the stack.
Under certain states, some features may not exist, and we use special tokens NULL$^w$, NULL$^t$ and NULL$^l$ to represent non-existent word, POS tag, and arc label features, respectively.
Our feature templates are similar to that of \newcite{chen-manning:2014:EMNLP2014}, except that we do not leverage features from the set, 
because the words inside the set are unordered.

\subsection{The light version}

We also consider a light version of our linearizer that only leverages words and unlabeled dependency relations.
Similar to Section \ref{sec:model}, the system also uses a feed-forward neural network with 1 hidden layer, but only takes word features as input.
It uses 4 types of actions: \textsc{Shift}-$w$, \textsc{LeftArc}, \textsc{RightArc}, and \textsc{End}.
All actions are same as described in Section \ref{sec:actions}, except that \textsc{LeftArc} and \textsc{RightArc} are not associated with arc labels.
Given a set of $n$ words as input, the system takes $2n$ steps to synthesize the sentence, which is faster and less vulnerable to error propagation.

\section{Integrating an LSTM language model}

Our model can be integrated with the baseline multi-layer LSTM language model.
Existing work \cite{zhang-blackwood-clark:2012:EACL2012,liu-zhang:2015:EMNLP} has shown that a syntactic linearizer can benefit from a surface language model by taking its scores as features.
Here we investigate two methods for the integration: (1) joint decoding by interpolating the conditional probabilities and (2) feature-level integration by taking the output vector ($\mathbf{h}_{I}$) of the LSTM language model as features to the linearizer.

\subsection{Joint decoding}
\label{sec:joint_decoding}

To perform joint decoding, the conditional action probability distributions of both models given the current state are interpolated, and the best action under the interpolated probability distribution is chosen, before both systems advancing to a new state using the action.
The interpolated conditional probability is:
\begin{multline}
p(a|s_i,h_i;\theta_1,\theta_2) = \log p(a|s_i; \theta_1) \\
+ \alpha \log p(a|h_i; \theta_2)\textrm{,}
\label{eq:comb}
\end{multline}
where $s_i$ and $\theta_1$ are the state and parameters of the linearizer, $h_i$ and $\theta_2$ are the state and parameters of the LSTM language model, and $\alpha$ is the interpolation hyper parameter.

The action spaces of the two systems are different because the actions of the LSTM language model correspond only to the shift actions of the linearizer.
To match the probability distributions, we expand the distribution of the LSTM language model as shown in Equation \ref{eq:comb_prob}, where $w_a$ is the associated word of a shift action $a$.
Generally, the probabilities of non-shift actions are 1.0, and those of shift actions are from the LSTM language model with respect to $w_a$:
\begin{equation}
p(a|h_i; \theta_2) = \begin{cases} p(w_a|h_i; \theta_2), & \mbox{if } a\mbox{ is shift} \\ 
1.0, & \mbox{otherwise} \end{cases}
\label{eq:comb_prob}
\end{equation}
We do not normalize the interpolated probability distribution,
because our experiments show that normalization only gives around 0.3 BLEU score gains, while significantly decreasing the speed.
When a shift action is chosen, both systems advance to a new state; otherwise only the linearizer advances to a new state.

\subsection{Feature level integration}

To take the output of an LSTM language model as a feature in our model, we first train the LSTM language model independently. 
During the training of our model, we take $\textbf{h}_{I}$, the output of the top LSTM layer after consuming all words on the stack, as a feature in the input layer of Figure \ref{fig:sys},
before finally advancing both the linearizer and the LSTM language model using the predicted action.
This is analogous to adding a separately-trained $n$-gram language model as a feature to a discriminative linearizer \cite{liu-zhang:2015:EMNLP}.
Compared with joint decoding (Section \ref{sec:joint_decoding}), $p(a|s_i,h_i;\theta_1,\theta_2)$ is calculated by one model, and thus there is no need to tune the hyper-parameter $\alpha$.
The state update remains the same: the language model advances to a new state only when a shift action is taken.

\section{Training}
\label{sec:training}

\begin{table*} \small
\centering
\begin{tabular}{lcccccccc}
\hline
System  & \multicolumn{2}{c}{\textsc{BeamSize=1}} & \multicolumn{2}{c}{\textsc{BeamSize=10}} & \multicolumn{2}{c}{\textsc{BeamSize=64}} & \multicolumn{2}{c}{\textsc{BeamSize=512}} \\
        & BLEU & Time & BLEU & Time & BLEU & Time & BLEU & Time \\
\hline
\hline
\textsc{LSTM} & 14.01 & 6m26s & 26.83 & 13m & 33.05 & 54m41s & 37.08 & 405m10s \\
\hline
\textsc{Syn}  & 20.97 & 11m39s & 27.72 & 26m40s & 30.01 & 113m19s & 31.12 & 891m39s \\
\textsc{Syn$+$LSTM} & 21.17 & 18m15s & 30.43 & 37m15s & 34.35 & 157m16s & 36.84 & 1058m \\
\textsc{Syn$\times$LSTM} & \textbf{24.91} & 18m12s & 32.75 & 37m12s & 35.88 & 156m50s & 36.96 & 1070m \\
\textsc{Syn$_{l}\times$LSTM} & 24.55 & 9m50s & \textbf{32.84} & 23m7s & \textbf{36.11} & 77m6s & \textbf{37.99} & 624m39s \\
\hline
\end{tabular}
\caption{Main results and decoding times.}
\label{tab:test}
\end{table*}

Following \newcite{chen-manning:2014:EMNLP2014}, we set the training objective as maximizing the log-likelihood of each successive action conditioned on the dependency tree, which can be gold or automatically parsed.
To train our linearizer, we first generate training examples $\{(s_i, t_i)\}_{i=1}^{m}$ from the training sentences and their gold parse trees, where $s_i$ is a state, and $t_i \in T$ is the corresponding oracle transition.
We use the ``arc standard'' oracle \cite{nivre2008algorithms}, which always prefers \textsc{Shift} over \textsc{leftArc}.
The final training objective is to minimize the
cross-entropy loss, plus an L2-regularization term:

\[
L(\theta)=-\sum_{i}\log p_{t_i} + \frac{\lambda}{2}\|\theta\|^2\textrm{,}
\]
where $\theta$ represents all the trainable parameters: ${\mathbf{W}_1, \mathbf{b}_1, \mathbf{W}_2, \mathbf{E}^w, \mathbf{E}^t, \mathbf{E}^l}$.
A slight variation is that the softmax probabilities are computed only among the feasible transitions in practice.
As described in Section \ref{sec:actions}, for an input set of words, the feasible transitions are: \textsc{Shift}-$w$, where $w$ is a word in the set, 
\textsc{Pos}-$p$ for all POS tags, \textsc{LeftArc}-$l$ and \textsc{RightArc}-$l$ for all arc labels, and \textsc{End}.

To train a linearizer that takes an LSTM language model as features, we first train the LSTM language model on the same training data, then train the linearizer with the parameters of the LSTM language model unchanged.

\begin{table} \small
\centering
\begin{tabular}{cccc}
\hline
ID & \#training sent & \#iter & F1 \\
\hline
\hline
syn90 & all  & 30 & 90.28 \\
syn85 & all  & 1  & 85.38 \\
syn79 & 9000 & 1  & 79.68 \\
syn54 & 900  & 1  & 54.86 \\
\hline
\end{tabular}
\caption{Parsing accuracy settings, the F1 scores are measured on the training set.}
\label{tab:pas}
\end{table}

\section{Experiments}

\subsection{Setup}

We follow previous work and conduct experiments on the Penn Treebank, using Wall Street Journal sections 2-21 for training, 22 for development and 23 for final testing.
Gold-standard dependency trees are derived from bracketed sentences in the treebank using Penn2Malt.\footnote{https://stp.lingfil.uu.se/$\sim$nivre/research/Penn2Malt.html}
In order to study the influence of parsing accuracy of the training data, we use ten-fold jackknifing to construct WSJ training data with different accuracies. 
More specifically, the data is first randomly split into ten equal-size subsets, and then each subset is automatically parsed with a constituent parser trained on the other subsets, before the results are finally converted to dependency trees using Penn2Malt.
In order to obtain datasets with different parsing accuracies, we randomly sample a small number of sentences from each training subset and choose different training iterations, as shown in Table \ref{tab:pas}.
In our experiments, we use ZPar\footnote{https://github.com/frcchang/zpar} \cite{zhu-EtAl:2013:ACL20131} for automatic constituent parsing.

Our syntactic linearizer is implemented with Keras.\footnote{https://keras.io/}
We randomly initialize $\mathbf{E}^w$, $\mathbf{E}^t$, $\mathbf{E}^l$, $\mathbf{W}^1$ and $\mathbf{W}^2$ within $(-0.01, 0.01)$, and use default setting for other parameters.
The hyper-parameters and parameters which achieve the best performance on the development set are chosen for final evaluation.
Our vocabulary comes from SENNA\footnote{http://ronan.collobert.com/senna/}, which has 130,000 words.
The activation functions $\mathrm{tanh}$ and $\mathrm{softmax}$ are added on top of the hidden and output layers, respectively.
We use Adagrad \cite{duchi2011adaptive} with an initial learning rate of 0.01, regularization parameter $\lambda=10^{-8}$, and dropout rate 0.3 for training.
The interpolation coefficient $\alpha$ for joint decoding is set 0.4.
During decoding, simple pruning methods are applied, such as a constraint that \textsc{Pos}-$p$ actions always follow \textsc{Shift}-$w$ actions.

We evaluate our linearizer (\textsc{Syn}) and its variances, where the subscript ``$_l$'' denotes the light version, ``$+$\textsc{LSTM}'' represents joint decoding with an LSTM language model, and ``$\times$\textsc{LSTM}'' represents taking an LSTM language model as features in our model.
We compare results with the current state-of-the-art: an LSTM (\textsc{LSTM}) language model from \newcite{schmaltz-rush-shieber:2016:EMNLP2016}, which is similar in size and architecture to the medium LSTM setup of \newcite{zaremba2014recurrent}.
None of the systems use future cost heuristic.
All experiments are conducted using Tesla K20Xm.

\subsection{Tuning}

We show some development results in this section.
First, using the cube activation function \cite{chen-manning:2014:EMNLP2014} does not yield a good performance on our task.
We tried other activations including $\mathrm{Linear}$, $\mathrm{tanh}$ and $\mathrm{ReLU}$ \cite{nair2010rectified}, and $\mathrm{tanh}$ gives the best results.
In addition, we tried pretrained embeddings from SENNA, which does not yield better results compared to random initialization.
Further, dropout rates from $0.3$ to $0.8$ give good training results.
Finally, we tried different values from $0.1$ to $1.0$ for the interpolation coefficient $\alpha$, finding that values between $0.3$ and $0.7$ give the best performances, while values larger than $1.5$ yield poor performances.

\subsection{Main results}

\begin{table*}[t] \small
\centering
\begin{tabularx}{\textwidth}{l|X} 
\hline
System & sentences \\
\hline
\hline
\textsc{LSTM}-512 & the bush administration , known as 31 , 1992 , earlier this year said it would extend voluntary restraint agreements steel quotas until march . \\
\textsc{Syn$_{l}\times$LSTM}-512 & earlier this year , the bush administration said it would extend steel agreements until march 31 , 1992 , known as voluntary restraint quotas . \\
\textsc{Ref} & the bush administration earlier this year said it would extend steel quotas , known as voluntary restraint agreements , until march 31 , 1992 . \\
\hline
\hline
\textsc{LSTM}-512 & shearson lehman hutton inc. said , however , that it is `` going to set back with the customers , '' because of friday 's plunge , president of jeffrey b. lane concern `` reinforces volatility relations . \\
\textsc{Syn$_{l}\times$LSTM}-512 & however , jeffrey b. lane , president of shearson lehman hutton inc. , said that friday 's plunge is `` going to set back with customers because it reinforces the volatility of `` concern , '' relations . \\
\textsc{Ref} & however , jeffrey b. lane , president of shearson lehman hutton inc. , said that friday 's plunge is `` going to set back '' relations with customers , `` because it reinforces the concern of volatility . \\
\hline
\hline
\textsc{LSTM}-512 & the debate between the stock and futures markets is prepared for wall street will cause another situation about whether de-linkage 
crash undoubtedly properly renewed friday . \\
\textsc{Syn$_{l}\times$LSTM}-512 & the wall street futures markets undoubtedly will cause renewed debate about whether the stock situation is properly prepared for an
other crash between friday and de-linkage . \\
\textsc{Ref} & the de-linkage between the stock and futures markets friday will undoubtedly cause renewed debate about whether wall street is prope
rly prepared for another crash situation . \\
\hline
\end{tabularx}
\caption{Output samples.}
\label{tab:example}
\end{table*}

The main results on the test set are shown in Table \ref{tab:test}.
Compared with previous work, our linearizers achieve the best results under all beam sizes,
especially under the greedy search scenario (\textsc{BeamSize}=1), where \textsc{Syn} and \textsc{Syn$\times$LSTM} outperform the baseline of \textsc{LSTM} by 7 and 11 BLEU points, respectively.
This demonstrates that syntactic information is extremely important when beam size is small.
In addition, our syntactic systems are still better than the baseline under very large beam sizes (such as, \textsc{BeamSize}=512), which lead to slow performance and are less useful practically.
On the other hand, the baseline (\textsc{LSTM}) benefits more from beam size increases.
The results are consistent with \cite{ma-zhang-zhu:2014:P14-2} in that both increasing beam size and using richer features are solutions for error propagation.

\textsc{Syn$\times$LSTM} is better than \textsc{Syn$+$LSTM}. 
In fact, \textsc{Syn$\times$LSTM} can be considered as interpolation with $\alpha$ being automatically calculated under different states.
Finally, \textsc{Syn$_{l}\times$LSTM} is better than \textsc{Syn$\times$LSTM} except under greedy search, showing that word-to-word dependency features may be sufficient for this task.

As for the decoding times, \textsc{Syn$_{l}\times$LSTM} shows a moderate time growth along increasing beam size, which is roughly 1.5 times slower than \textsc{LSTM}.
In addition, \textsc{Syn$+$LSTM} and \textsc{Syn$\times$LSTM} are the slowest for each beam size (roughly 3 times slower than \textsc{LSTM}), because of the large number of features they use and the large number of decoding steps they take.
\textsc{Syn} is roughly 2 times slower than \textsc{LSTM}.

Previous work, such as \newcite{schmaltz-rush-shieber:2016:EMNLP2016}, adopts future cost and the information of base noun phrase (BNP) and shows further improvement on performance.
However, these are highly task specific.
Future cost is based on the assumption that all words are available at the beginning, which is not true for other tasks. 
On the other hand, our model does not rely on this assumption, thus can be better applicable on other tasks.
BNPs are the phrases that correspond to leaf NP nodes in constituent trees. 
Assuming BNPs being available is not practical either.

\begin{figure}[t]
\centering
\includegraphics[scale=0.4]{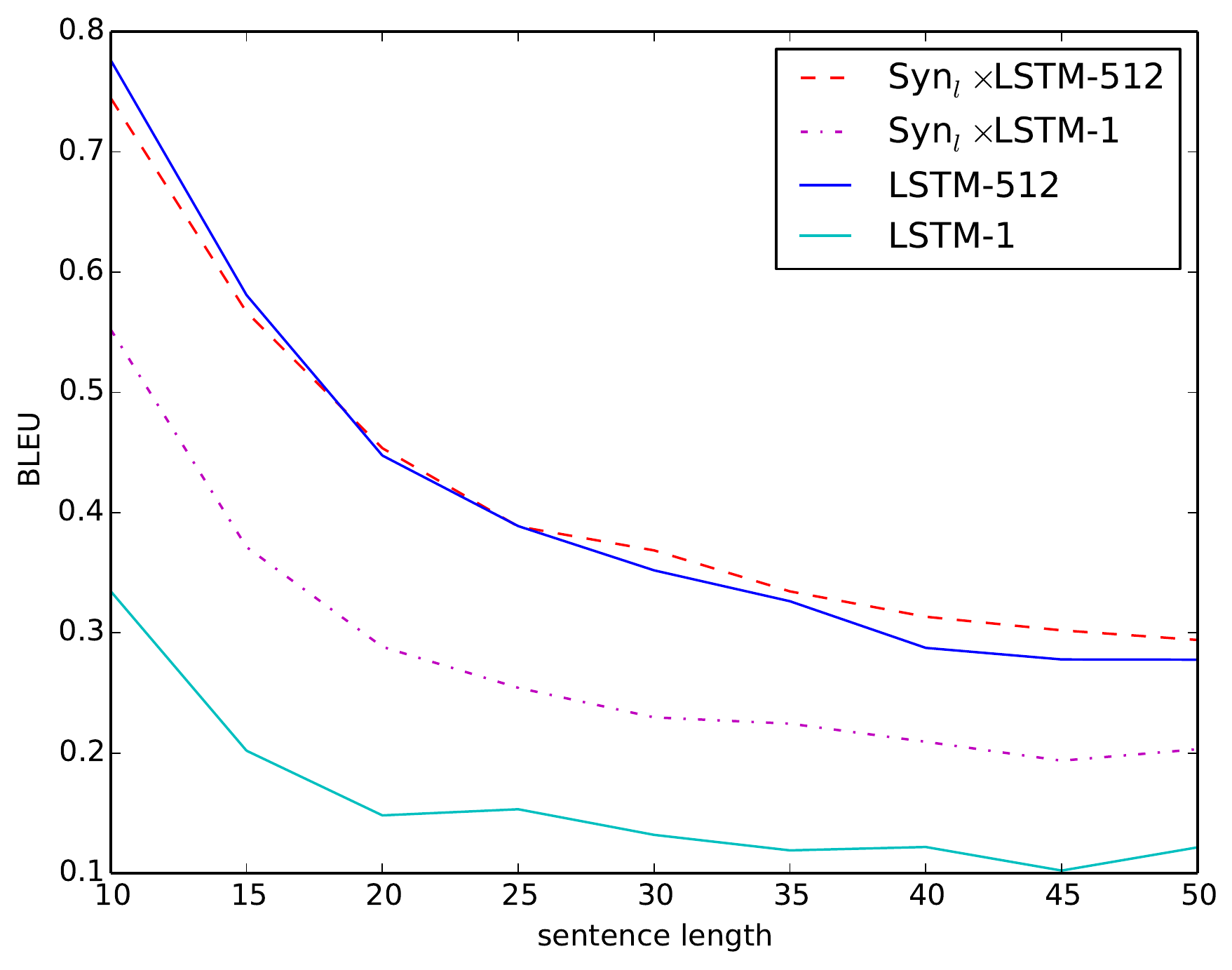}
\caption{Performance on different lengths.}
\label{fig:bleu_len}
\end{figure}


\subsection{Influence of sentence length}

We show the performances on different sentence lengths in Figure \ref{fig:bleu_len}.
The results are from \textsc{LSTM} and \textsc{Syn$_{l}\times$LSTM} using beam size 1 and 512.
Sentences belonging to the same length range (such as 1--10 or 11--15) are grouped together, and corpus BLEU is calculated on each group.
First of all, \textsc{Syn$_{l}\times$LSTM-1} is significantly better than \textsc{LSTM-1} on all sentence lengths, explaining the usefulness of syntactic features.
In addition, \textsc{Syn$_{l}\times$LSTM-512} is notably better than \textsc{LSTM-512} on sentences that are longer than 25, and the difference is even larger on sentences that have more than 35 words.
This is an evidence that \textsc{Syn$_{l}\times$LSTM} is better at modeling long-distance dependencies.
On the other hand, \textsc{LSTM-512} is better than \textsc{Syn$_{l}\times$LSTM-512} on short sentences (length $\le$ 10).
The reason may be that \textsc{LSTM} is good at modeling relatively shorter dependencies without syntactic guidance, while \textsc{Syn$_{l}\times$LSTM}, which takes more steps for synthesizing the same sentence, suffers from error propagation.
Overall, this figure can be regarded as empirical evidence that syntactic systems are better choices for generating long sentences \cite{wan-EtAl:2009:EACL,zhang-clark:2011:EMNLP}, while surface systems may be better choices for generating short sentences.

Table \ref{tab:example} shows some linearization results of long sentences from \textsc{LSTM} and \textsc{Syn$_{l}\times$LSTM} using beam size 512.
The outputs of \textsc{Syn$_{l}\times$LSTM} are notably more grammatical than those of \textsc{LSTM}.
For example, in the last group, the output of \textsc{Syn$_{l}\times$LSTM} means ``the market will cause another debate about whether the situation now is prepared for another crash'', while the output of \textsc{LSTM} is obviously less fluent, especially for the parts ``... markets is prepared for wall street will cause ...'' and ``... crash undoubtedly properly renewed ..''.

In addition, \textsc{LSTM} makes locally grammatical outputs, while suffering more mistakes in the global level.
Taking the second group as an example, \textsc{LSTM} generates grammatical phrases, such as ``going to set back with the customers'' and ``because of friday 's plunge'', while misplacing ``president of'', which should be in the very front of the sentence.
On the other hand, \textsc{Syn$_{l}\times$LSTM} can capture patterns such as ``president of some inc.'' and ``someone, president of someplace said'' to make the right choices.
Finally, \textsc{Syn$_{l}\times$LSTM} can makes grammatical sentences with different meanings.
For example in the first group, the result of \textsc{Syn$_{l}\times$LSTM} means ``the bush administration will extend the steel agreement'', while the true meaning is ``the bush administration will extend the steel quotas''.
For syntactic linearization, such semantic variation is tolerable.

\subsection{Results with auto-parsed data}

\begin{table} \small
\centering
\begin{tabular}{l|c|c}
\hline
Data & \textsc{Syn$\times$LSTM} & \textsc{Syn$_l\times$LSTM} \\
\hline
\hline
Gold     & 36.03 & 36.41 \\
syn90    & 35.91 & 36.31 \\
syn85    & 35.84 & 36.22 \\
syn79    & 35.40 & 35.96 \\
syn54    & 33.32 & 34.98 \\
\hline
\end{tabular}
\caption{Results of various parsing accuracy.}
\label{tab:auto_res}
\end{table}

There is no syntactically annotated data in many domains.
As a result, performing syntactic linearization in these domains requires automatically parsed training data, which may affect the performance of our syntactic linearizer.
We study this effect by training both \textsc{Syn$\times$LSTM} and \textsc{Syn$_{l}\times$LSTM} with automatically parsed training data of different parsing accuracies, and show the results, which are generated with beamsize 64 on the devset, in Table \ref{tab:auto_res}.
Generally, a higher parsing accuracy can lead to a better linearization result for both systems.
It conforms to the intuition that syntactic quality affects the fluency of surface texts.
On the other hand, the influence is not large, the BLEU scores of \textsc{Syn$_{l}\times$LSTM} and \textsc{Syn$\times$LSTM} drop by 1.5 and 2.8 BLEU points, respectively, as the parsing accuracy decreases from gold to 54\%.
Both observations are consistent with that of \newcite{liu-zhang:2015:EMNLP} for discrete syntactic linearization.
Finally, \textsc{Syn$_{l}\times$LSTM} shows less BLEU score decreases than \textsc{Syn$\times$LSTM}.
The reason is that \textsc{Syn$_{l}\times$LSTM} only takes word features, and is less vulnerable to parsing accuracy decrease.

\subsection{Embedding similarity}

\begin{table} \small
\centering
\begin{tabular}{l|l}
\hline
Actions & Top similar actions \\
\hline
\hline
S-wednesday & S-tuesday S-friday S-thursday S-monday \\
S-huge      & S-strong S-serious S-good S-large \\
S-taxes     & S-bills S-expenses S-loans S-payments \\
S-secretary & S-department S-officials S-director \\
S-largely   & S-partly S-primarily S-mostly S-entirely \\
\hline
\end{tabular}
\caption{Top similar actions for shift actions}
\label{tab:shift}
\end{table}

\begin{figure}
\centering
\includegraphics[scale=0.4]{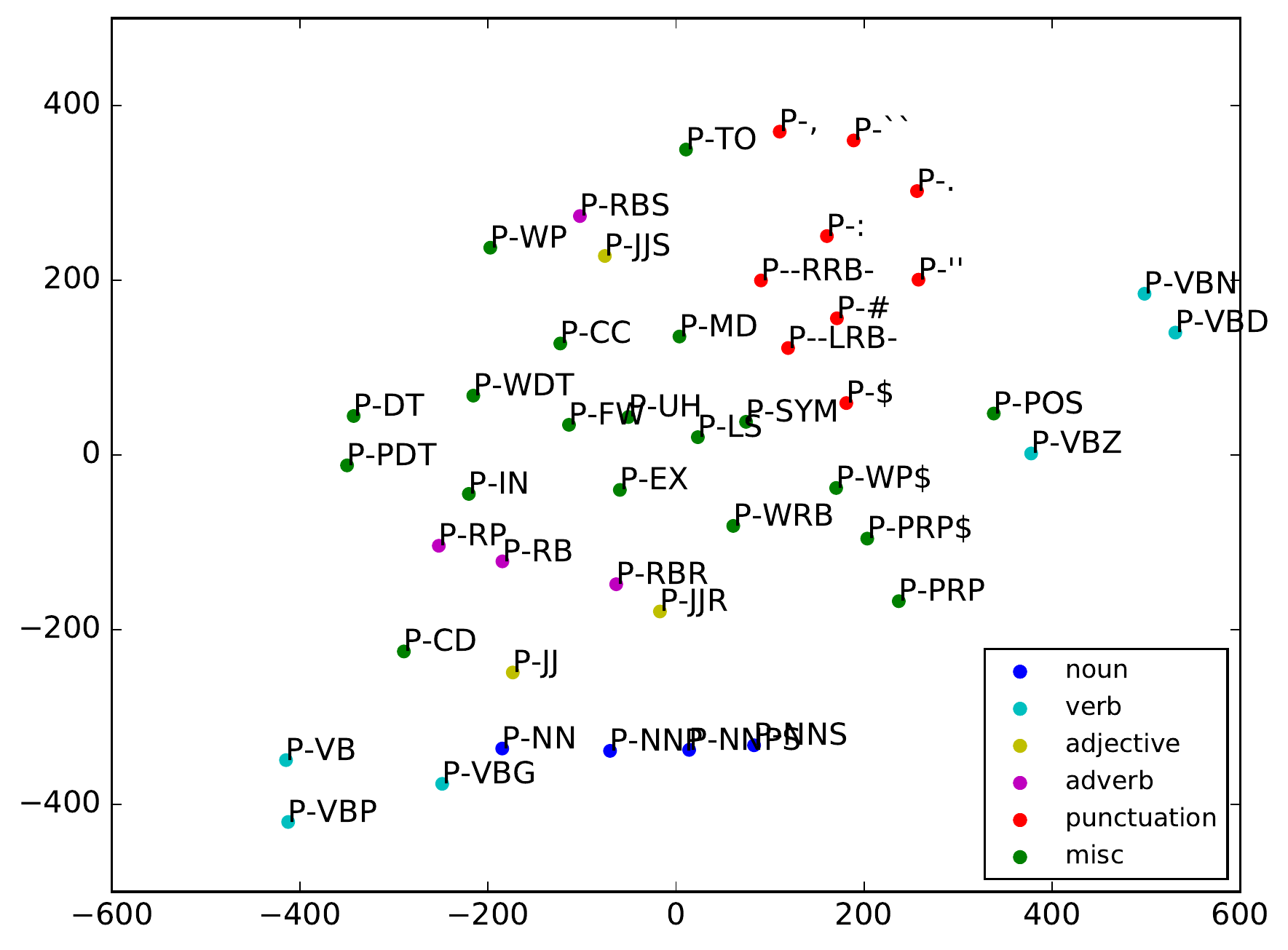}
\caption{t-SNE visualization of POS embeddings}
\label{fig:pos}
\end{figure}

One main advantage of neural systems is that they use vectorized features, which are less sparse than discriminative features.
Taking $\mathbf{W}_2$ as the embedding matrix of actions, we calculate the top similar actions for the \textsc{Shift}-$w$ actions by cosine distance and show examples in Table \ref{tab:shift}. 
In addition, Figure \ref{fig:pos} presents the t-SNE visualization \cite{maaten2008visualizing} of the embeddings for the \textsc{POS}-$p$ actions.
Generally, the embeddings of similar actions are closer than these of other actions.
From both results, we can see that our model learns reasonable embeddings from the Penn Treebank, a small-scale corpus, which shows the effectiveness of our system from another perspective.

\section{Conclusion}

We studied neural transition-based syntactic linearization, which combines the advantages of both neural networks and syntactic information.
In addition, we compared two ways of integrating a neural language model into our system.
Experimental results show that our system achieves improved results comparing with a state-of-the-art multi-layer LSTM language model.
To our knowledge, we are the first to investigate neural syntactic linearization.

In the future work, we will investigate LSTM on this task.
In particular, an LSTM decoder, taking features form the already-built subtrees as part of its inputs, is taken to model the sequences of shift-reduce actions.
Another possible direction is creating complete graphs with their nodes being the input words, before encoding them with self-attention networks \cite{vaswani2017attention} or graph neural networks \cite{kipf2016semi,P18-1026,P18-1030,P18-1150}.
This approach can be better at capturing word-to-word dependencies than simply summing word embeddings up.

\bibliography{emnlp2017}

\begin{thebibliography}{}
\expandafter\ifx\csname natexlab\endcsname\relax\def\natexlab#1{#1}\fi

\bibitem[{Beck et~al.(2018)Beck, Haffari, and Cohn}]{P18-1026}
Daniel Beck, Gholamreza Haffari, and Trevor Cohn. 2018.
\newblock Graph-to-sequence learning using gated graph neural networks.
\newblock In {\em Proceedings of the 56th Annual Meeting of the Association for
  Computational Linguistics (ACL-18)\/}.

\bibitem[{Bohnet et~al.(2010)Bohnet, Wanner, Mill, and
  Burga}]{bohnet-EtAl:2010:PAPERS}
Bernd Bohnet, Leo Wanner, Simon Mill, and Alicia Burga. 2010.
\newblock Broad coverage multilingual deep sentence generation with a
  stochastic multi-level realizer.
\newblock In {\em Proceedings of the 23rd International Conference on
  Computational Linguistics (COLING-10)\/}. Beijing, China, pages 98--106.

\bibitem[{Chen and Manning(2014)}]{chen-manning:2014:EMNLP2014}
Danqi Chen and Christopher Manning. 2014.
\newblock A fast and accurate dependency parser using neural networks.
\newblock In {\em Conference on Empirical Methods in Natural Language
  Processing (EMNLP-14)\/}. Doha, Qatar, pages 740--750.

\bibitem[{de~Gispert et~al.(2014)de~Gispert, Tomalin, and
  Byrne}]{degispert-tomalin-byrne:2014:EACL}
Adri\`{a} de~Gispert, Marcus Tomalin, and Bill Byrne. 2014.
\newblock Word ordering with phrase-based grammars.
\newblock In {\em Proceedings of the 14th Conference of the European Chapter of
  the ACL (EACL-14)\/}. Gothenburg, Sweden, pages 259--268.

\bibitem[{Duchi et~al.(2011)Duchi, Hazan, and Singer}]{duchi2011adaptive}
John Duchi, Elad Hazan, and Yoram Singer. 2011.
\newblock Adaptive subgradient methods for online learning and stochastic
  optimization.
\newblock {\em Journal of Machine Learning Research\/} 12(Jul):2121--2159.

\bibitem[{He et~al.(2009)He, Wang, Guo, and Liu}]{he-EtAl:2009:ACLIJCNLP}
Wei He, Haifeng Wang, Yuqing Guo, and Ting Liu. 2009.
\newblock Dependency based chinese sentence realization.
\newblock In {\em Proceedings of the 47th Annual Meeting of the Association for
  Computational Linguistics (ACL-09)\/}. Suntec, Singapore, pages 809--816.

\bibitem[{Kiddon et~al.(2016)Kiddon, Zettlemoyer, and
  Choi}]{kiddon-zettlemoyer-choi:2016:EMNLP2016}
Chlo\'{e} Kiddon, Luke Zettlemoyer, and Yejin Choi. 2016.
\newblock Globally coherent text generation with neural checklist models.
\newblock In {\em Conference on Empirical Methods in Natural Language
  Processing (EMNLP-16)\/}. Austin, Texas, pages 329--339.

\bibitem[{Kipf and Welling(2016)}]{kipf2016semi}
Thomas~N Kipf and Max Welling. 2016.
\newblock Semi-supervised classification with graph convolutional networks.
\newblock {\em arXiv preprint arXiv:1609.02907\/} .

\bibitem[{Liu and Zhang(2015)}]{liu-zhang:2015:EMNLP}
Jiangming Liu and Yue Zhang. 2015.
\newblock An empirical comparison between n-gram and syntactic language models
  for word ordering.
\newblock In {\em Conference on Empirical Methods in Natural Language
  Processing (EMNLP-15)\/}. Lisbon, Portugal, pages 369--378.

\bibitem[{Liu et~al.(2015)Liu, Zhang, Che, and Qin}]{liu-EtAl:2015:NAACL-HLT1}
Yijia Liu, Yue Zhang, Wanxiang Che, and Bing Qin. 2015.
\newblock Transition-based syntactic linearization.
\newblock In {\em Conference on Empirical Methods in Natural Language
  Processing (EMNLP-15)\/}. Denver, Colorado, pages 113--122.

\bibitem[{Ma et~al.(2014)Ma, Zhang, and Zhu}]{ma-zhang-zhu:2014:P14-2}
Ji~Ma, Yue Zhang, and Jingbo Zhu. 2014.
\newblock Punctuation processing for projective dependency parsing.
\newblock In {\em Proceedings of the 52nd Annual Meeting of the Association for
  Computational Linguistics (ACL-14)\/}. Baltimore, Maryland, pages 791--796.

\bibitem[{Maaten and Hinton(2008)}]{maaten2008visualizing}
Laurens van~der Maaten and Geoffrey Hinton. 2008.
\newblock Visualizing data using t-sne.
\newblock {\em Journal of Machine Learning Research\/} 9(Nov):2579--2605.

\bibitem[{Nair and Hinton(2010)}]{nair2010rectified}
Vinod Nair and Geoffrey~E Hinton. 2010.
\newblock Rectified linear units improve restricted {B}oltzmann machines.
\newblock In {\em Proceedings of the 27th international conference on machine
  learning (ICML-10)\/}. pages 807--814.

\bibitem[{Nivre(2008)}]{nivre2008algorithms}
Joakim Nivre. 2008.
\newblock Algorithms for deterministic incremental dependency parsing.
\newblock {\em Computational Linguistics\/} 34(4):513--553.

\bibitem[{Papineni et~al.(2002)Papineni, Roukos, Ward, and
  Zhu}]{papineni-EtAl:2002:ACL}
Kishore Papineni, Salim Roukos, Todd Ward, and Wei-Jing Zhu. 2002.
\newblock Bleu: a method for automatic evaluation of machine translation.
\newblock In {\em Proceedings of the 40th Annual Conference of the Association
  for Computational Linguistics (ACL-02)\/}. Philadelphia, Pennsylvania, USA,
  pages 311--318.

\bibitem[{Schmaltz et~al.(2016)Schmaltz, Rush, and
  Shieber}]{schmaltz-rush-shieber:2016:EMNLP2016}
Allen Schmaltz, Alexander~M. Rush, and Stuart Shieber. 2016.
\newblock Word ordering without syntax.
\newblock In {\em Conference on Empirical Methods in Natural Language
  Processing (EMNLP-16)\/}. Austin, Texas, pages 2319--2324.

\bibitem[{Serban et~al.(2016)Serban, Garc\'{i}a-Dur\'{a}n, Gulcehre, Ahn,
  Chandar, Courville, and Bengio}]{serban-EtAl:2016:P16-1}
Iulian~Vlad Serban, Alberto Garc\'{i}a-Dur\'{a}n, Caglar Gulcehre, Sungjin Ahn,
  Sarath Chandar, Aaron Courville, and Yoshua Bengio. 2016.
\newblock Generating factoid questions with recurrent neural networks: The 30m
  factoid question-answer corpus.
\newblock In {\em Proceedings of the 54th Annual Meeting of the Association for
  Computational Linguistics (ACL-16)\/}. Berlin, Germany, pages 588--598.

\bibitem[{Song et~al.(2014)Song, Zhang, Song, and Liu}]{song2014joint}
Linfeng Song, Yue Zhang, Kai Song, and Qun Liu. 2014.
\newblock Joint morphological generation and syntactic linearization.
\newblock In {\em Proceedings of the National Conference on Artificial
  Intelligence (AAAI-14)\/}. pages 1522--1528.

\bibitem[{Song et~al.(2018)Song, Zhang, Wang, and Gildea}]{P18-1150}
Linfeng Song, Yue Zhang, Zhiguo Wang, and Daniel Gildea. 2018.
\newblock A graph-to-sequence model for amr-to-text generation.
\newblock In {\em Proceedings of the 56th Annual Meeting of the Association for
  Computational Linguistics (ACL-18)\/}.

\bibitem[{Vaswani et~al.(2017)Vaswani, Shazeer, Parmar, Uszkoreit, Jones,
  Gomez, Kaiser, and Polosukhin}]{vaswani2017attention}
Ashish Vaswani, Noam Shazeer, Niki Parmar, Jakob Uszkoreit, Llion Jones,
  Aidan~N Gomez, {\L}ukasz Kaiser, and Illia Polosukhin. 2017.
\newblock Attention is all you need.
\newblock In {\em Advances in Neural Information Processing Systems\/}. pages
  5998--6008.

\bibitem[{Wan et~al.(2009)Wan, Dras, Dale, and Paris}]{wan-EtAl:2009:EACL}
Stephen Wan, Mark Dras, Robert Dale, and C\'{e}cile Paris. 2009.
\newblock Improving grammaticality in statistical sentence generation:
  Introducing a dependency spanning tree algorithm with an argument
  satisfaction model.
\newblock In {\em Proceedings of the 12th Conference of the European Chapter of
  the ACL (EACL-09)\/}. Athens, Greece, pages 852--860.

\bibitem[{Wen et~al.(2015)Wen, Gasic, Mrk\v{s}i\'{c}, Su, Vandyke, and
  Young}]{wen-EtAl:2015:EMNLP}
Tsung-Hsien Wen, Milica Gasic, Nikola Mrk\v{s}i\'{c}, Pei-Hao Su, David
  Vandyke, and Steve Young. 2015.
\newblock Semantically conditioned {LSTM}-based natural language generation for
  spoken dialogue systems.
\newblock In {\em Conference on Empirical Methods in Natural Language
  Processing (EMNLP-15)\/}. Lisbon, Portugal, pages 1711--1721.

\bibitem[{White(2005)}]{white:2005:Software}
Michael White. 2005.
\newblock Designing an extensible api for integrating language modeling and
  realization.
\newblock In {\em Proceedings of the ACL Workshop on Software\/}. Ann Arbor,
  Michigan, pages 47--64.

\bibitem[{White and Rajkumar(2009)}]{white-rajkumar:2009:EMNLP}
Michael White and Rajakrishnan Rajkumar. 2009.
\newblock Perceptron reranking for {CCG} realization.
\newblock In {\em Conference on Empirical Methods in Natural Language
  Processing (EMNLP-09)\/}. Singapore, pages 410--419.

\bibitem[{Zaremba et~al.(2014)Zaremba, Sutskever, and
  Vinyals}]{zaremba2014recurrent}
Wojciech Zaremba, Ilya Sutskever, and Oriol Vinyals. 2014.
\newblock Recurrent neural network regularization.
\newblock {\em arXiv preprint arXiv:1409.2329\/} .

\bibitem[{Zhang(2013)}]{zhang2013partial}
Yue Zhang. 2013.
\newblock Partial-tree linearization: Generalized word ordering for text
  synthesis.
\newblock In {\em Proceedings of the International Joint Conference on
  Artificial Intelligence (IJCAI-13)\/}.

\bibitem[{Zhang et~al.(2012)Zhang, Blackwood, and
  Clark}]{zhang-blackwood-clark:2012:EACL2012}
Yue Zhang, Graeme Blackwood, and Stephen Clark. 2012.
\newblock Syntax-based word ordering incorporating a large-scale language
  model.
\newblock In {\em Proceedings of the 13th Conference of the European Chapter of
  the ACL (EACL-12)\/}. Avignon, France, pages 736--746.

\bibitem[{Zhang and Clark(2011)}]{zhang-clark:2011:EMNLP}
Yue Zhang and Stephen Clark. 2011.
\newblock Syntax-based grammaticality improvement using {CCG} and guided
  search.
\newblock In {\em Conference on Empirical Methods in Natural Language
  Processing (EMNLP-11)\/}. Edinburgh, Scotland, UK., pages 1147--1157.

\bibitem[{Zhang et~al.(2018)Zhang, Liu, and Song}]{P18-1030}
Yue Zhang, Qi~Liu, and Linfeng Song. 2018.
\newblock Sentence-state lstm for text representation.
\newblock In {\em Proceedings of the 56th Annual Meeting of the Association for
  Computational Linguistics (ACL-18)\/}.

\bibitem[{Zhang and Nivre(2011)}]{zhang-nivre:2011:ACL-HLT2011}
Yue Zhang and Joakim Nivre. 2011.
\newblock Transition-based dependency parsing with rich non-local features.
\newblock In {\em Proceedings of the 49th Annual Meeting of the Association for
  Computational Linguistics (ACL-11)\/}. Portland, Oregon, USA, pages 188--193.

\bibitem[{Zhang et~al.(2014)Zhang, Song, Song, Zhu, and
  Liu}]{zhang-EtAl:2014:EMNLP20141}
Yue Zhang, Kai Song, Linfeng Song, Jingbo Zhu, and Qun Liu. 2014.
\newblock Syntactic {SMT} using a discriminative text generation model.
\newblock In {\em Conference on Empirical Methods in Natural Language
  Processing (EMNLP-14)\/}. Doha, Qatar, pages 177--182.

\bibitem[{Zhu et~al.(2013)Zhu, Zhang, Chen, Zhang, and
  Zhu}]{zhu-EtAl:2013:ACL20131}
Muhua Zhu, Yue Zhang, Wenliang Chen, Min Zhang, and Jingbo Zhu. 2013.
\newblock Fast and accurate shift-reduce constituent parsing.
\newblock In {\em Proceedings of the 51st Annual Meeting of the Association for
  Computational Linguistics (ACL-13)\/}. Sofia, Bulgaria, pages 434--443.

\end{thebibliography}
\bibliographystyle{emnlp_natbib}

\end{document}